\documentclass{article}

\usepackage[final]{infer2control_2018}

\usepackage{natbib}
\setcitestyle{square,sort,comma,numbers}
\usepackage{microtype}
\usepackage{graphicx}
\usepackage{subfigure}
\usepackage{booktabs} 
\usepackage{wrapfig}

\usepackage{hyperref}

\usepackage[utf8]{inputenc} 
\usepackage[T1]{fontenc}    
\usepackage{url}            
\usepackage{amsfonts}       
\usepackage{nicefrac}       
\usepackage{subfigure}
\graphicspath{{figures/}}
\usepackage{xcolor}
\usepackage{adjustbox}
\usepackage{bm}
\usepackage{amsmath}
\usepackage{amssymb}
\usepackage{algorithm}
\usepackage{algorithmic}
\usepackage{multirow}
\usepackage{adjustbox}

\newcommand{\RN}[1]{%
	\textup{\lowercase\expandafter{\it \romannumeral#1}}%
}

\title{Scalable Thompson Sampling via Optimal Transport} 

\input{Definitions.tex}
\author{
   Ruiyi Zhang$^1$,  Zheng Wen$^2$, Changyou Chen$^3$ and Lawrence Carin$^1$\\
   {$^1$Duke University,~$^2$Adobe Research,~$^3$SUNY at Buffalo}\\
   \texttt{ryzhang@cs.duke.edu, zheng.wen@adobe.com}
}

\begin{document}

\maketitle
\vspace{-5mm}
\begin{abstract}
\vspace{-3mm}
  Thompson sampling (TS) is a class of algorithms for sequential decision-making, which requires maintaining a posterior distribution over a model. However, calculating exact posterior distributions is intractable for all but the simplest models. Consequently, efficient computation of an approximate posterior distribution is a crucial problem for scalable TS with complex models, such as neural networks. In this paper, we use distribution optimization techniques to approximate the posterior distribution, solved via Wasserstein gradient flows. Based on the framework, a principled particle-optimization algorithm is developed for TS to approximate the posterior efficiently. Our approach is scalable and does not make explicit distribution assumptions on posterior approximations.  Extensive experiments on both synthetic data and real large-scale data demonstrate the superior performance of the proposed methods.	  
\end{abstract}

\vspace{-6mm}
\section{Introduction}
\vspace{-2mm}
In many online sequential decision-making problems, such as contextual bandits \citep{bubeck2012regret} and reinforcement learning \citep{sutton1998introduction}, an agent needs to learn to take a sequence of actions to maximize its expected cumulative reward, while repeatedly interacting with an unknown environment. Moreover, since in such problems the agent's actions affect both its rewards and its observations, it faces the well-known exploration-exploitation dilemma. Consequently, the exploration strategy is crucial for a learning algorithm: typically, under-exploration will make the algorithm stick at a sub-optimal strategy, while over-exploration tends to incur a huge exploration cost.

Various exploration strategies have been proposed, including $\epsilon$-greedy (EG), Boltzmann exploration~\citep{sutton1990integrated, cesa2017boltzmann}, upper-confidence bound (UCB)~\citep{agrawal1995sample, auer2002using} type exploration, and Thompson sampling (TS). 
Among them, TS \citep{thompson1933likelihood, russo2018tutorial}, which is also known as posterior sampling or probability matching, is a widely used exploration strategy with good practical performance~\citep{li2010contextual, chapelle2011empirical} and theoretical guarantees~\citep{russo2016information, agrawal2012analysis, russo2018tutorial}. 
 The vanilla TS is computationally efficient when there is a closed-form posterior, such as with Bernoulli or Gaussian rewards. For cases without a closed-form posterior, other variants of TS have also been developed~\citep{agrawal2013thompson,chapelle2011empirical}. However, most such TS algorithms cannot be extended to cases with complex generalization models, such as neural networks, in a computationally efficient manner.
%
%

In this paper, adopting ideas from Wasserstein-gradient-flow literature, we propose a general particle-based distribution optimization framework for Thompson sampling. Our framework improves the recently proposed particle-based framework for TS \citep{lu2017ensemble}. Specifically, our framework employs a set of particles interacting with each other to approximate the posterior distribution (thus it is called particle-interactive Thompson sampling, or $\pi$-TS), while \citep{lu2017ensemble} treats particles independently.
%
Specifically, we optimize the posterior distribution in TS based on the Wasserstein-gradient-flow framework. In this setting, Bayesian sampling in TS becomes a convex optimization problem on the space of {\em probability measures}, thus the optimality of the learned distribution could be guaranteed. For tractability, the posterior distribution in TS is approximated by a set of particles (a.k.a.\! samples). We test our framework on a number of applications, first in a simulated dynamic scenario and then on large-scale real-world datasets. In all cases, our proposed particle-interactive Thompson sampling significantly outperforms other baselines.
\vspace{-3mm}
\section{Background}
\vspace{-2mm}
\subsection{Contextual Bandits and Thompson Sampling}
\label{sec:contextual_bandit}
We consider a contextual bandit characterized by a triple $\left(\mathcal{X}, \mathcal{A}, P\right)$, where $\mathcal{X} \subseteq \Re^d$ is a context (state) space with dimension $d$, $\mathcal{A} = \{1, 2, \dots , K\}$ is a finite action space, and $P$ encodes the reward distributions at all the context-action pairs. 
The agent is assumed to know $\mathcal{X}$ and $\mathcal{A}$, but not $P$. The agent repeatedly interacts with the contextual bandit for $T$ rounds. At each round $t=1, \ldots, T$, the agent first observes a context $\xb_t \in \mathcal{X}$, which is independently chosen by the \emph{environment}. Then, the agent adaptively chooses an action $\ab_t \in \mathcal{A}$, based on the current context $\xb_t$ and the agent's past observations. Finally, the agent observes and receives a reward $r_t$, which is conditionally independently drawn from the reward distribution $P_{\xb_t, \ab_t}$. 
%
The agent's objective is to learn 
 to minimize its expected cumulative regret in the first $T$ rounds, {\it i.e.}, $R(T)= \sum_{t=1}^T \mathbb{E} \left[ \max_{\ab\in\mathcal{A}}[{r}(\xb_t, \ab)] - r_t\right]$.

Many practical online decision-making problems that fit into the framework of contextual bandit have intractably large scale. Specifically, in such problems, at least one of $\mathcal{X}$ and $\mathcal{A}$ has unmanageably large cardinality (if it is discrete) or dimension (if it is continuous). One standard approach to develop scalable learning algorithms for such large-scale problems is to exploit generalization models.
Specifically, in this paper, we assume that the learning agent has access to a generalization model $m(\xb, \ab; \thetab)$ for the mean reward function $\bar{r}(\xb, \ab)$, where $\thetab$ is the model parameters. We assume that the generalization model $m$ is ``accurate'' in the sense that there exists a specific model parameter vector $\thetab^*$, with
$
m(\xb, \ab; \thetab^*) \approx \bar{r}(\xb, \ab), ~\forall (\xb, \ab)
\in \mathcal{X} \times \mathcal{A} .$
Note that $m$ is a function of the context-action pair $(\xb, \ab)$ and the parameter vector $\thetab$. 
%
%
One example of the above-mentioned generalization model is neural network and simpler generalization models, such as linear regression models and logistic regression models, can be viewed as special cases of it.


Thompson sampling (TS)~\citep{thompson1933likelihood} is a widely used class of algorithms for sequential decision-making. For the contextual bandits with reward generalization, assuming the reward generalization is perfect, the vanilla version of TS proceeds as follows: 
Given a prior distribution $p_0(\bm{\theta})$ on the model parameters $\bm{\theta}$ and the set of past observations $\mathcal{D}_t \triangleq \{(\mathbf{x}_i, \ab_i, r_{i})\}_{i=1}^{t-1}$,
Thompson sampling~\citep{thompson1933likelihood} maintains a posterior distribution over $\thetab$ as $p_{t-1} \triangleq p(\thetab|\Dcal_{t-1})$.
%
%
Then, at each time $t$, it first samples a parameter vector $\hat{\thetab}_t$ from the current posterior $p_{t-1}$; then, it chooses action $\ab_t \in \argmax_{\ab} m(\xb_t, \ab; \hat{\thetab}_t)$, and receives the reward $r_t$; finally it updates the posterior. The pseudocode is provided in Algorithm~\ref{algo:vanilla}.



\subsection{Wasserstein Gradient Flows}
Wasserstein gradient flows (WGF) is a generalization of gradient flows on Euclidean space (reviewed in Section \ref{sec:gf_euc}), by lifting the differential equation above onto the space of probability measures, denoted $\mathcal{P}(\Omega)$ with $\Omega \subset \mathbb{R}^d$. Formally, we first endow a Riemannian geometry~\citep{carmo1992riemannian} on $\mathcal{P}(\Omega)$.
The geometry is characterized by the length between two elements (two distributions), defined by the second-order Wasserstein distance:
{\small$\label{eq:w2dist}\nonumber
	W_2^2(\mu, \nu) \triangleq \inf_{\gamma}\left\{\int_{\Omega \times \Omega}\|\thetab - \thetab^\prime\|_2^2\mathrm{d}\gamma(\thetab, \thetab^\prime): \gamma \in \Gamma(\mu, \nu)\right\}
	$},
where $\Gamma(\mu, \nu)$ is the set of joint distributions over $(\thetab, \thetab^\prime)$ such that the two marginals equal $\mu$ and $\nu$, respectively. 
%
%
If $\mu$ is absolutely continuous w.r.t.\! the Lebesgue measure, there is a unique optimal transport plan from $\mu$ to $\nu$, {\it i.e.}, a mapping $T: \mathbb{R}^d \rightarrow\mathbb{R}^d$ pushing $\mu$ onto $\nu$ satisfying $T_{\#}\mu = \nu$. Here $T_{\#}\mu$ denotes the pushforward measure~\citep{Villani:08} of $\mu$.
The Wasserstein distance thus can be equivalently reformulated as
{\small$
	W_2^2(\mu, \nu) \triangleq \inf_{T}\left\{\int_{\Omega}\|\thetab - T(\thetab) \|^2_2\mathrm{d}\mu(\thetab)\right\}$}. 

Consider $\mathcal{P}(\Omega)$ with a Riemannian geometry endowed by the second-order Wasserstein metric.
Let $\{\mu_{\tau}\}_{\tau\in[0,1]}$ be an absolutely continuous curve in $\mathcal{P}(\Omega)$ with distance between $\mu_{\tau}$ and $\mu_{\tau+h}$ measured by $W_2^2(\mu_{\tau}, \mu_{\tau+h})$. We overload the definition of $T$ to denote the underlying transformation from $\mu_{\tau}$ to $\mu_{\tau+h}$ as $\thetab_{\tau+h} = T_h(\thetab_{\tau})$. Motivated by the Euclidean-space case, if we define $\vb_{\tau}(\thetab) \triangleq \lim_{h\rightarrow 0}\frac{T_h(\thetab_{\tau}) - \thetab_{\tau}}{h}$ as the {\em velocity of the particle}, a gradient flow can be defined on $\mathcal{P}(\Omega)$ correspondingly in Lemma \ref{theo:gf_w_exist}~\citep{Ambrosio:book05}.
\vspace{-0.2cm}
\begin{lemma}\label{theo:gf_w_exist}
	Let $\{\mu_{\tau}\}_{\tau\in[0,1]}$ be an absolutely-continuous curve in $\mathcal{P}(\Omega)$ with finite second-order 
	moments. Then for a.e.\! $\tau\in [0, 1]$, the above vector field $\vb_{\tau}$ defines a gradient flow on $\mathcal{P}(\Omega)$ as $\partial_{\tau} \mu_{\tau} + \nabla_{\thetab} \cdot (\vb_{\tau} \mu_{\tau}) = 0$, where $\nabla_{\thetab}\cdot\ab \triangleq \nabla_{\thetab}^{\top} \ab$ for a vector $\ab$.
\end{lemma}
\vspace{-2mm}
Function $F$ is in the space of probability measures $\mathcal{P}(\Omega)$, mapping a probability measure $\mu$ to a real value, {\it i.e.}, $F: \mathcal{P}(\Omega) \rightarrow \mathbb{R}$.  Consequently, it can be shown that $\vb_{\tau}$ in Lemma~\ref{theo:gf_w_exist} has the form $\vb_{\tau} = -\nabla_{\xb} \frac{\delta F}{\delta \mu_{\tau}}(\mu_{\tau})$ \citep{Ambrosio:book05}, where $\frac{\delta F}{\delta \mu_{\tau}}$ is called the {\em first variation of functional} $F$ at $\mu_{\tau}$~\citep{dougan2012first}.
Based on this, gradient flows on $\mathcal{P}(\Omega)$ can be written in a form of partial differential equation (PDE) as
\vspace{-0.15cm}
{\small
	\begin{equation}
	\begin{aligned}\label{eq:gf_dis1}
	\partial_{\tau} \mu_{\tau} = -\nabla_{\thetab} \cdot (\vb_{\tau} \mu_{\tau}) = \nabla_{\thetab} \cdot \left(\mu_{\tau} \nabla_{\thetab}(\frac{\delta F}{\delta \mu_{\tau}}(\mu_{\tau}))\right)~.
	\end{aligned}
	\end{equation}
}
\vspace{-8mm}
\section{Thompson Sampling via Optimal Transport}
This section describes our proposed Particle-Interactive Thompson sampling ($\pi$-TS) framework. We first interpret Thompson sampling as a WGF problem, then propose an energy function to design a specific WGF, and finally propose particle-approximation methods to solve the $\pi$-TS problem.
The posterior distribution of $\thetab$ in Thompson sampling is defined as $p_{t-1} \triangleq p(\thetab|\Dcal_{t-1}) \propto e^{U(\thetab)}$, where the potential energy is defined as
\vspace{-2.5mm}
{
\small
\begin{align}
U(\thetab)\triangleq \log p(\mathcal{D}|\thetab)+\log p_0(\thetab)
= \sum_{i=1}^t\left(\log p(r_i|\xb_i, \ab_i, \thetab)+\frac{1}{t}\log p_0(\thetab)\right)
= \sum_{i=1}^NU_i(\thetab).
\end{align}}
\!\!To apply WGFs for posterior approximation in Thompson sampling, a variational (posterior) distribution for $\thetab$, denoted as $\mu(\thetab)$, is learned by solving an appropriate gradient-flow problem. To make the stationary distribution of the WGF consistent with the target posterior distribution, we define an energy functional characterizing the similarity between the current variational distribution and the true distribution $p_t$ induced by the rewards as:
\vspace{-2.5mm}
\begin{equation}
\small
\begin{aligned}\label{eq:imp_energy}
F(\mu) \triangleq -{\int U(\thetab)\mu(\thetab)\mathrm{d}\thetab} +
{\int \mu(\thetab)\log\mu(\thetab)\mathrm{d}\thetab}
= \KL\left(\mu\|p_{t}\right)~.
\end{aligned}
\end{equation}
%
The energy functional $F(\mu)$ defines a landscape determined by the rewards, whose minimum is obtained at $\mu = p_{t}$. We investigate the discrete-gradient-flow (DGF) method to solve \eqref{eq:gf_dis1}.

\begin{wrapfigure}[13]{R}{0.45\textwidth}
	\begin{minipage}[T]{0.43\textwidth}
		\vspace{-10mm}
		\begin{algorithm}[H]
			\caption{Particle-Interactive TS ($\pi$-TS)}
			\label{algo:algo2}
			\begin{algorithmic}[1]
				\REQUIRE {$\mathcal{D}_0 = \emptyset$; initialize particles $ \Theta_0  = \{\thetab_0^{i}\}_{i=1}^M$};
				\FOR{$t = 1,2,\ldots,T$}
				\STATE Observe context $\xb_t$
				\STATE Draw $\hat{\thetab}_{t}$ uniformly from $\Theta_{t}$
				\STATE Select $a_t \in \arg\max_{\ab} m(\xb_t, \ab; \hat{\thetab}_{t})$
				\STATE Observe and receive reward $r_t$
				\STATE $D_{t+1} = D_t\cup(\mathbf{x}_{t} , \mathbf{a}_{t}, r_{t})$
				\STATE Update $\Theta_{t+1}$, according to (\ref{eq:svgd_update3})
				\ENDFOR
			\end{algorithmic}
		\end{algorithm}
	\end{minipage}
\end{wrapfigure}

Discrete gradient flows (DGFs) approximate \eqref{eq:gf_dis1} by discretizing the continuous curve $\mu_t$ into a piece-wise linear curve, leading to an iterative optimization problem to solve the intermediate points denoted as $\{\mu_k^{h}\}_k$, where $k$ denotes the discrete points, and $h$ is refered to as the stepsize parameter. The iterative optimization problem is also known as Jordan-Kinderleher-Otto (JKO) scheme \citep{jordan1998variational}, where for iteration $k$, $\mu_{k+1}^{(h)}$ is obtained by solving the following optimization problem:
\vspace{-2mm}
{\small
\begin{align}\label{eq:ito_discrete}
\mu_{k+1}^{(h)} = \arg\min_{\mu}\KL\left(\mu\|p_\thetab\right) + \frac{W_2^2(\mu, \mu_k^{(h)})}{2h}~.
\end{align}
}
\!\!Following methods such as those in \citep{chen2018unified}, we  proposed to use particle approximation to approximate $\mu$ with $M$ particles $\{\thetab^{i}\}_{i=1}^M$ as 
$\mu^{(h)} \approx \frac{1}{M}\sum_{i=1}^M\delta_{\thetab^{i}}$,
where $\delta_{\thetab_k}$ is a delta function with a spike at $\thetab_k$. Consequently, the evolution of distributions described by \eqref{eq:gf_dis1} can be approximated with gradient ascent on particles. Specifically, \eqref{eq:ito_discrete} can be decomposed as 
%
$F_1 \triangleq -\mathbb{E}_{\mu}[\log p(\thetab|\mathcal{D})] + \lambda_1\mathbb{E}_{\mu}[\log \mu], 
F_2 \triangleq \lambda_2\mathbb{E}_{\mu}[\log \mu] + \frac{1}{2h}W_2^2(\mu, \mu_k^{(h)})$.
According to \cite{liu2016stein}, the gradient of the first term can be easily approximated as, $
{\small {\partial{F_1}}/{\partial{\thetab^{i}_k}} =  \sum_{j=1}^M \left[ -\kappa({\thetab}_{k}^{j}, {\thetab}_{k}^{i}) \nabla_{\thetab_{k}^{j}} U( {\thetab}_{k}^{i}) + \nabla_{{\thetab}_{k}^{j}} \kappa({\thetab}_{k}^{j}, {\thetab}_{k}^{i})\right]}
$, where $\kappa$ is the kernel function, which typically is the RBF kernel defined as $\kappa(\thetab, \thetab') = \exp(-\|\thetab - \thetab'\|^2_2/h)$.
For the second term $F_2$, we can solve the entropy-regularized Wasserstein distance
by introducing Lagrangian multipliers as:
$\label{eq:gradf2}
{\partial{F_2}}/{\partial{\thetab^{i}_k}}\approx -\frac{\sum_ju_iv_jc_{ij}e^{-\frac{c_{ij}}{\lambda_2}}}{\partial \thetab^{i}_k} 
= \sum_j 2u_iv_j\left(\frac{c_{ij}}{\lambda_2} - 1\right)
\exp^{  -\frac{c_{ij}}{\lambda_2}  }(\thetab^i_k-\thetab^j_{k-1})$.
where $c_{ij} \triangleq \|\thetab^i - \thetab^j\|_2^2$. Theoretically, we need to adaptively update Lagrangian multipliers $\{u_i, v_j\}$ as well to ensure the constraints in \eqref{eq:gradf2}. In practice, however, we use a fixed scaling factor $\gamma$ to approximate $u_iv_j$ for the sake of simplicity. The entropy-regularized Wasserstein term $F_2$ works as a complex force between particles in two ways: $\RN{1})$ When $\frac{c_{ij}}{\lambda} > 1$, $\thetab_k^{i}$ is pulled close to previous particles $\{\thetab_{k-1}^{j}\}$, with force proportional to $(\frac{c_{ij}}{\lambda} - 1)e^{-c_{ij}/\lambda}$; $\RN{2})$ when $\thetab^{i}$ is close enough to a previous particle $\thetab_k^{j}$, {\it i.e.}, $\frac{c_{ij}}{\lambda} < 1$, $\thetab_k^{i}$ is pushed away, preventing it from collapsing to $\thetab_k^{j}$.
Formally, in the $k$-th iteration, the particles are updated with:
\vspace{-2mm}
{\small
	\begin{align} \label{eq:svgd_update3}
	{\thetab}_{k+1}^{i} = {\thetab}_{k}^{i} + \dfrac{h}{M}\sum_{j=1}^M \left[ -\kappa({\thetab}_{k}^{j}, {\thetab}_{k}^{i}) \nabla_{{\thetab}_{k}^{i}} U( {\thetab}_{k}^{i}) + \nabla_{{\thetab}_{k}^{j}} \kappa({\thetab}_{k}^{j}, {\thetab}_{k}^{i}) + 
	\left(\frac{c_{ij}}{\lambda_2} - 1\right) \exp^{  -\frac{c_{ij}}{\lambda_2}  }(\thetab^i_k-\thetab^j_{k-1})\right]
	\end{align}
	\vspace{-4mm}
}

By applying the methods above to solve the WGF for Thompson sampling, we arrive at the Particle-Interactive Thompson Sampling ($\pi$-TS) framework. The pseudocode of $\pi$-TS is described in Algorithm \ref{algo:algo2}. In $\pi$-TS, the initial particles are drawn from the model prior $p_0(\thetab)$, which are maintained updated iteratively via discrete gradient flow to approximate the posterior distributions. Different from vanilla Thompson sampling, one approximate posterior sample is randomly selected from the particle set $\Theta_{t-1}$ in each iteration of $\pi$-TS to make decisions at time t.

\vspace{-3mm}
\section{Experiments}
\vspace{-2mm}
We conduct experiments to verify the performance of our proposed $\pi$-TS framework in both static scenarios and contextual-bandit problems. Our implementation is in TensorFlow and will be released upon publication. All computations were run on a single Tesla P100 GPU and all results are averaged over 50 realizations. 
\vspace{-3mm}
\subsection{Linear and Sparse Linear Contextual Bandits}
\vspace{-2mm}
\label{sec:exp_linearcase}
We consider a contextual bandit scenario where uncertainty estimation is driven by sequential decision making. This is more challenging because in this case the observations $\mathcal{D}$ are no longer i.i.d., leading to larger accumulative error as time goes on.
We test the proposed method in the linear setting~\citep{riquelme2018deep}. 
%
We can see from the figure that the proposed method, $\pi$-TS-DGF, performs almost as well as Lin-TS, the exact model; whereas other methods such as Neural-Linear and VI-TS receive much larger regrets. The gap is mostly caused by the approximation error between the exact posterior and approximate posterior. Especially, VI-TS shows a higher regret variance. 
\vspace{-3mm}
\subsection{Deep Contextual Bandits}
\begin{figure*}[ht]\centering
	\vspace{-3mm}
	\includegraphics[width=1.0\linewidth]{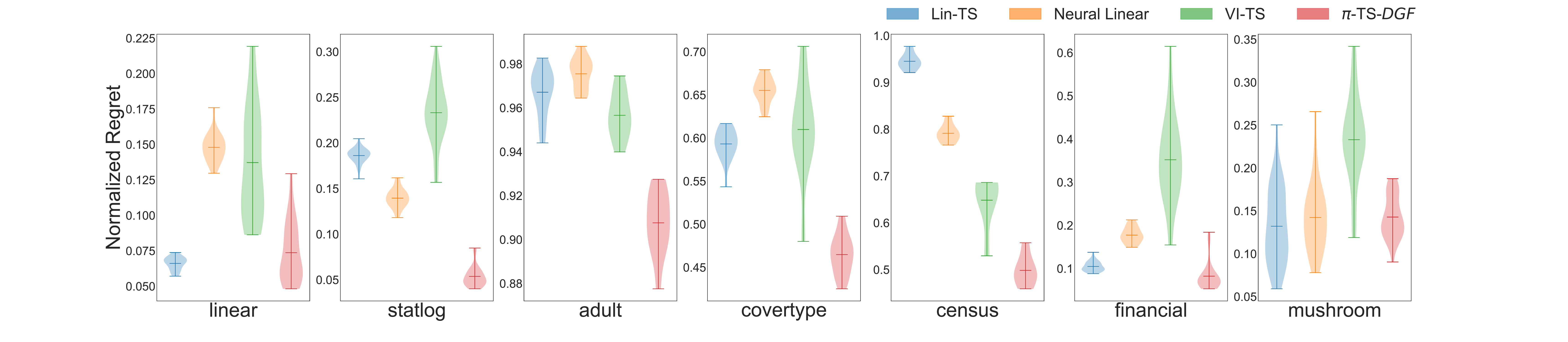}
	\vspace{-6mm}
	\caption{Normalized Cumulative Regret on all settings.}
	\label{fig:nregret}
	\vspace{-4mm}
\end{figure*}

\vspace{-2mm}
\label{sec:exp_deep}
Following the settings of \citep{riquelme2018deep}, we evaluate the algorithms on a range of bandit problems created from real-world data: Statlog, Covertype, Adult, Census, Financial, and Mushroom datasets. We normalize the cumulative regrets relative to that of the Uniform action selection, and plot the box-plot of the final normalized regrets in Figure \ref{fig:nregret}. In Figure \ref{fig:deep} are shown the mean (dark curves) and standard derivation (light areas) of regrets, along with number of pulls over 50 realizations.

To conclude, $\pi$-TS outperforms other methods; the performance of Lin-TS is not as good due to its poor representation.
With more data observed, it becomes increasingly difficult to approximate the exact posterior with the Lin-TS. With feature extracted by a neural network, the Neural Linear improves the performance and generally outperforms Lin-TS. Nevertheless, there are some cases where valid features cannot be well extracted by neural networks, leading to poor performance of Neural Linear. Furthermore, VI-TS consistently performs poorly with very high variances. The main cause might be that the underestimated uncertainty would lead to poor exploration. Our proposed $\pi$-TS outperforms other methods, since it can provide better uncertainty estimation than VI-TS, and endows more representation power than Lin-TS. Importantly, the performance of $\pi$-TS for relatively large datasets is much better than that of other methods.

\vspace{-2mm}
\section{Conclusion}
\vspace{-2mm}
In this paper, we proposed a scalable Thompson sampling framework $\pi$-TS for posterior application in Thompson sampling. We approximate the posterior distribution without an explicit-form variational distribution assumption, which leverages more powerful uncertainty estimation ability. Importantly, our methods can be applied on large-scale problems with complex models, such as neural networks. Specifically, $\pi$-TS approximates a distribution by defining gradient flows on the space of probability measures, and uses particles for approximation. Extensive experiments are conducted, demonstrating the effectiveness and efficiency of our proposed $\pi$-TS framework. Interesting future work includes designing more practically efficient variants of $\pi$-TS, and developing theory to study general regret bounds of the algorithms, as was done in \citep{lu2017ensemble,zhang2018stochastic}.
\newpage

\bibliography{piTSv5}
\bibliographystyle{plain}

\appendix
\title{Supplementary Material \\Scalable Thompson Sampling via Optimal Transport} 
\maketitle
\newpage

\section{Details of Experiments}
\subsection{Brief Dataset Description}
The dimensions of actions and contexts of different datasets are shown in Table \ref{tab:dataset}.
\begin{table}[h]
	\centering
	\caption{\small
		Description of datasets
	}
	\begin{tabular}{lcc}
		\toprule[1.2pt]
		{Dataset}& Contexts & Actions\\
		\midrule
		Mushroom & 22 & 2\\
		Statlog & 16 & 7\\
		Covertype & 54 & 7\\
		Financial & 21 & 8\\
		Census & 389 & 9 \\
		Adult & 94 & 14\\
		\bottomrule[1.2pt]
	\end{tabular}
	\label{tab:dataset}
	\vspace{-3mm}
\end{table}

\subsection{More Details of the Results}
\begin{figure}[ht] \centering
	\vspace{-3mm}
	\begin{tabular}{ccc}
		\hspace{-4mm}
		\includegraphics[width=0.3\linewidth]{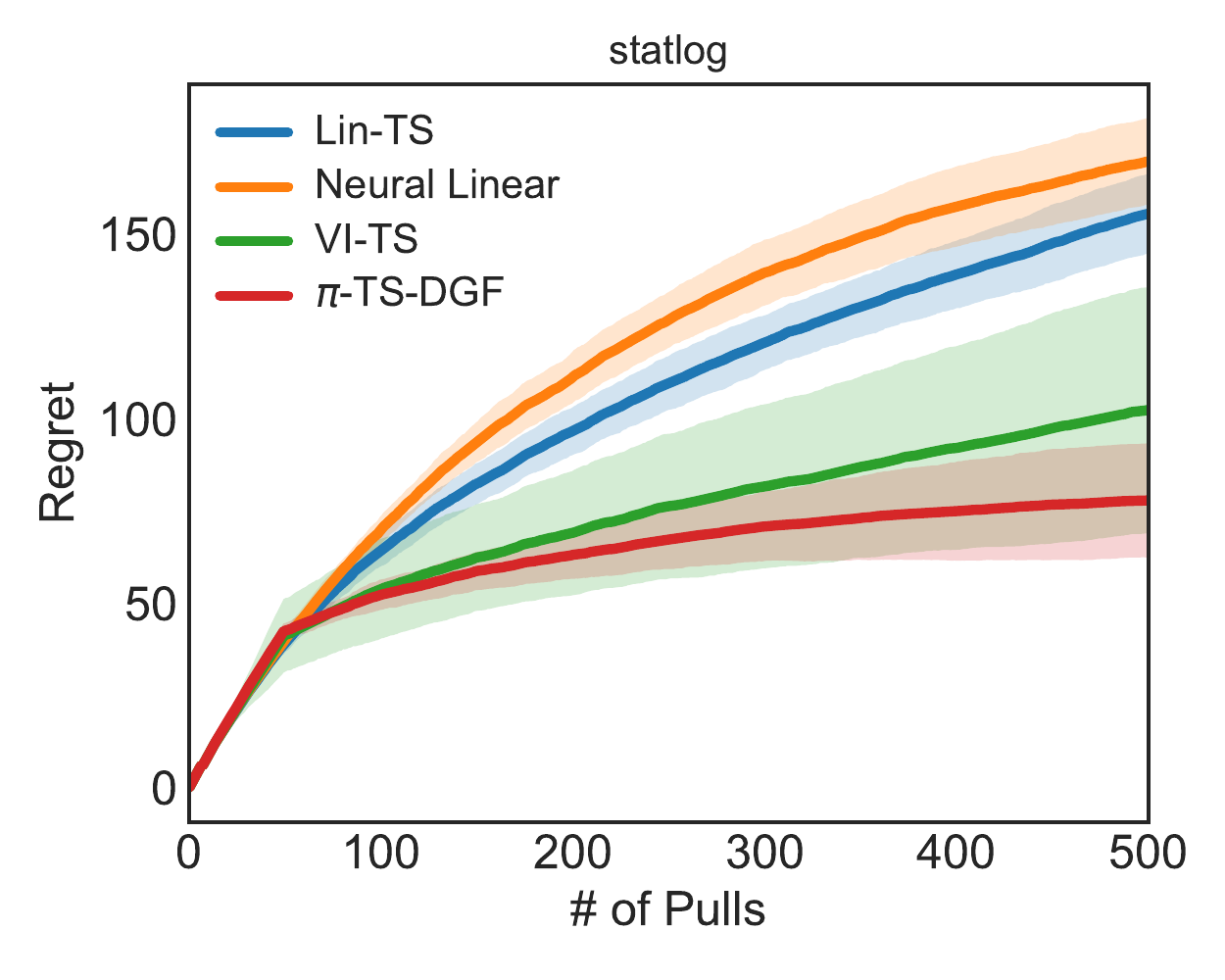}
		&   \hspace{-6mm}
		\includegraphics[width=0.3\linewidth]{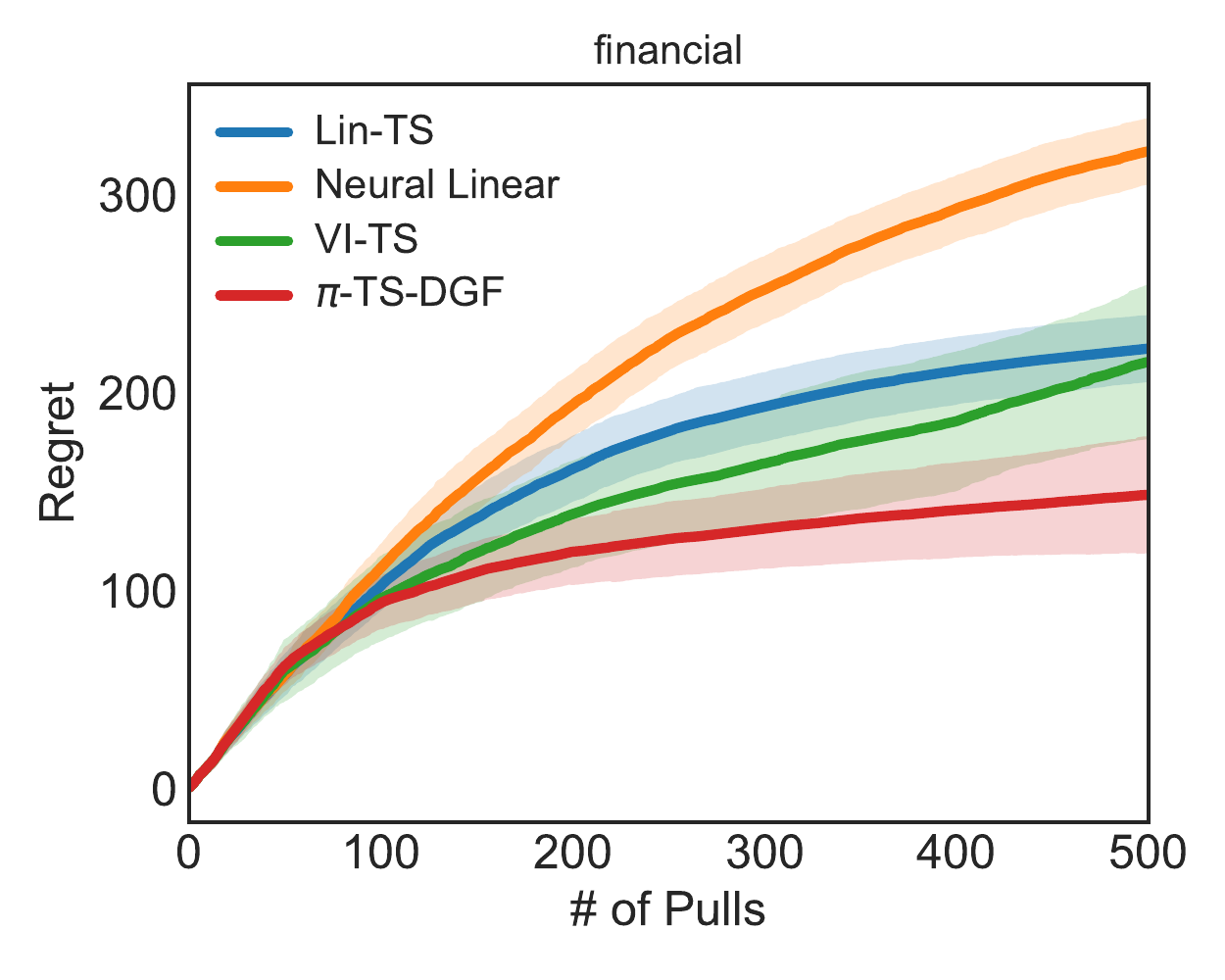}
		&   \hspace{-6mm}
		\includegraphics[width=0.3\linewidth]{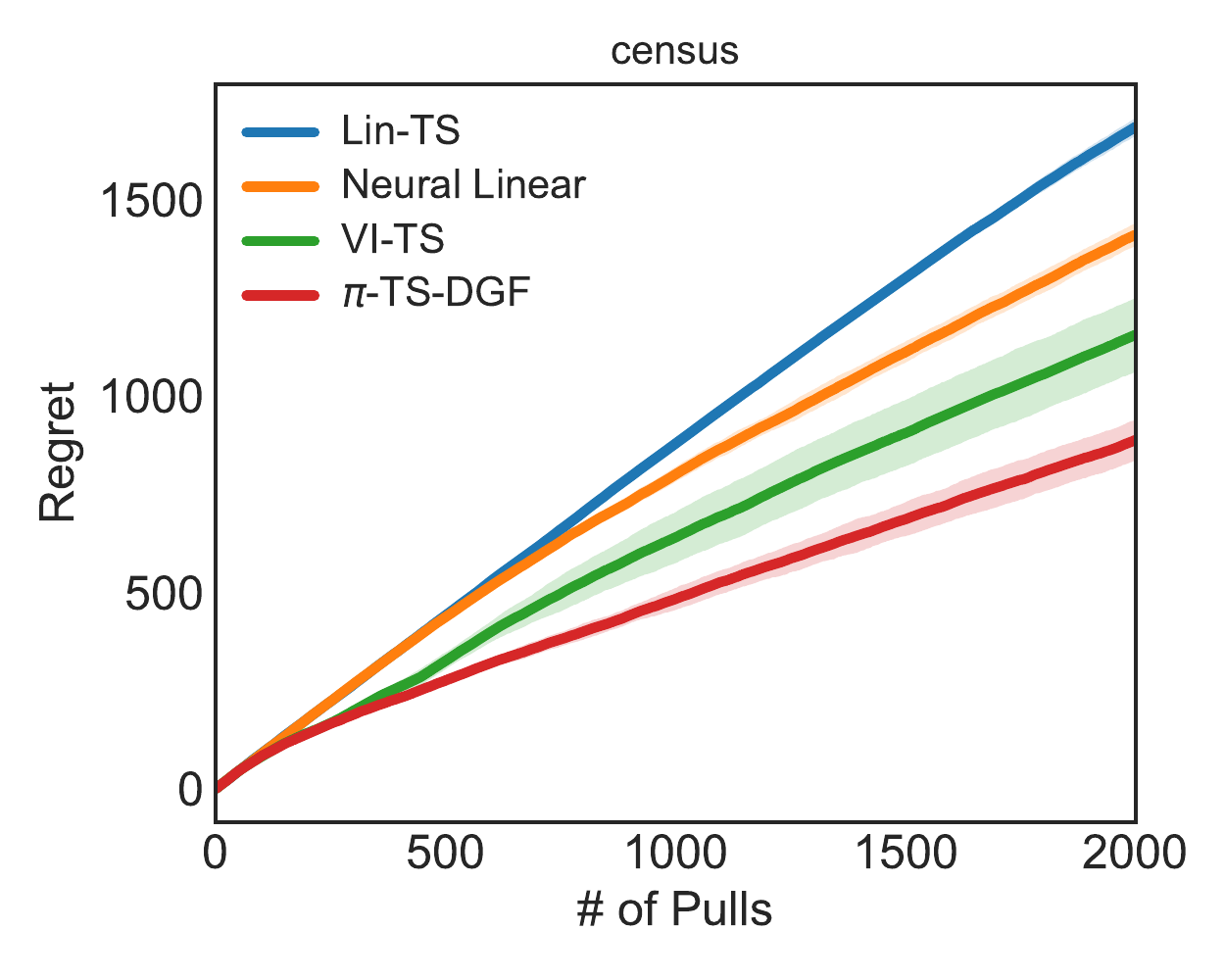}
		\vspace{-2mm}
		\\
		\hspace{-4mm}
		\includegraphics[width=0.3\linewidth]{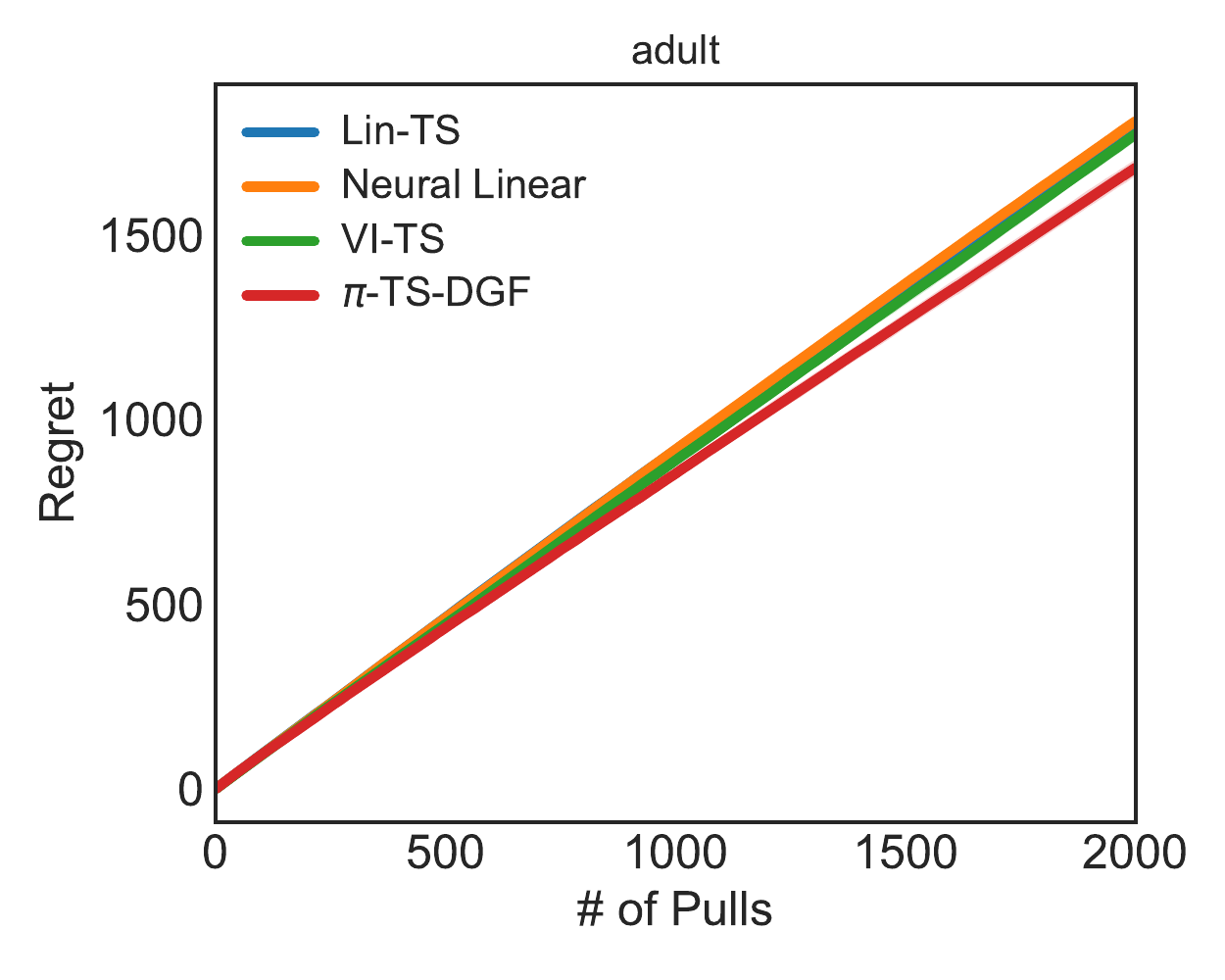}
		&   \hspace{-6mm}
		\includegraphics[width=0.3\linewidth]{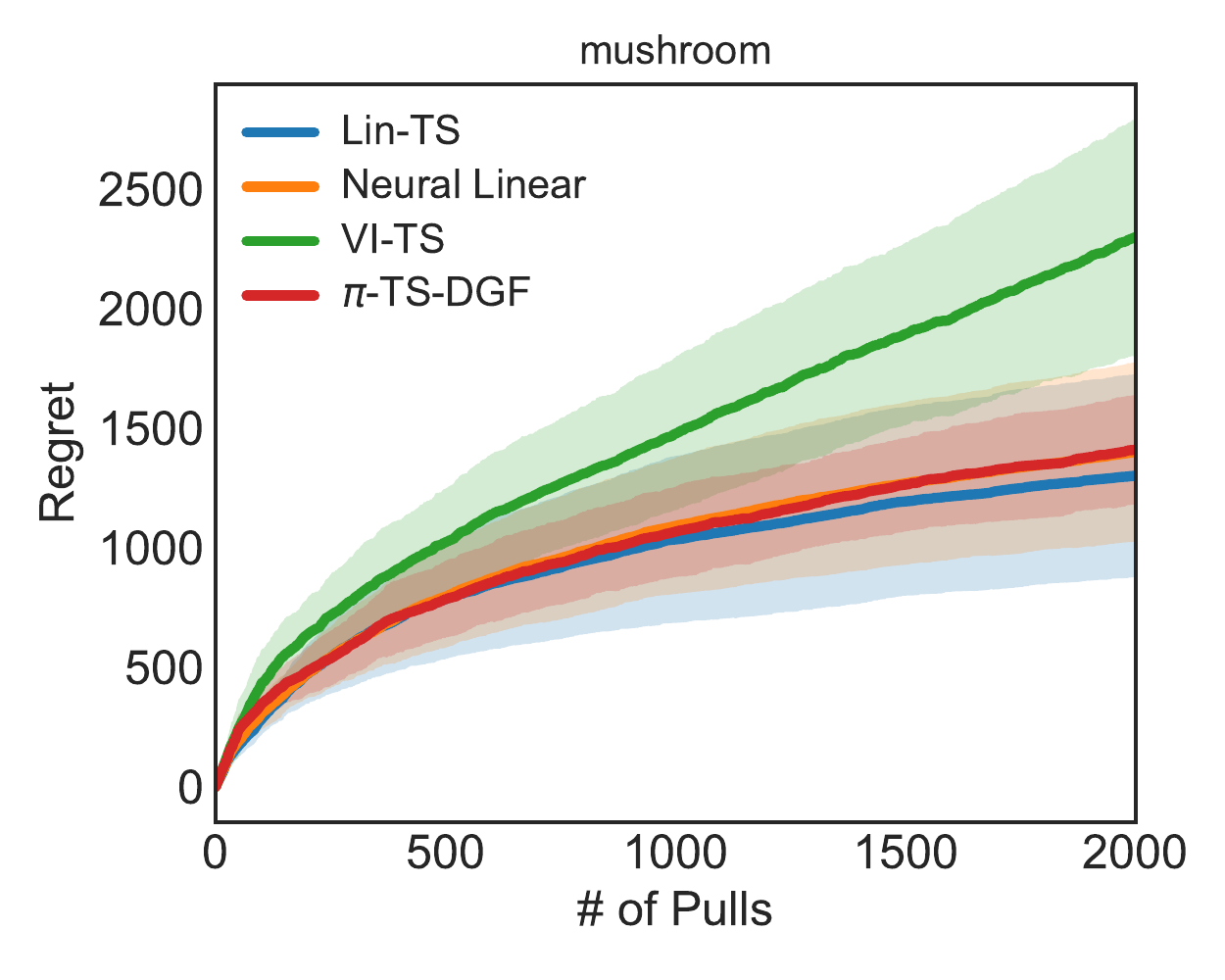}
		&   \hspace{-6mm}
		\includegraphics[width=0.3\linewidth]{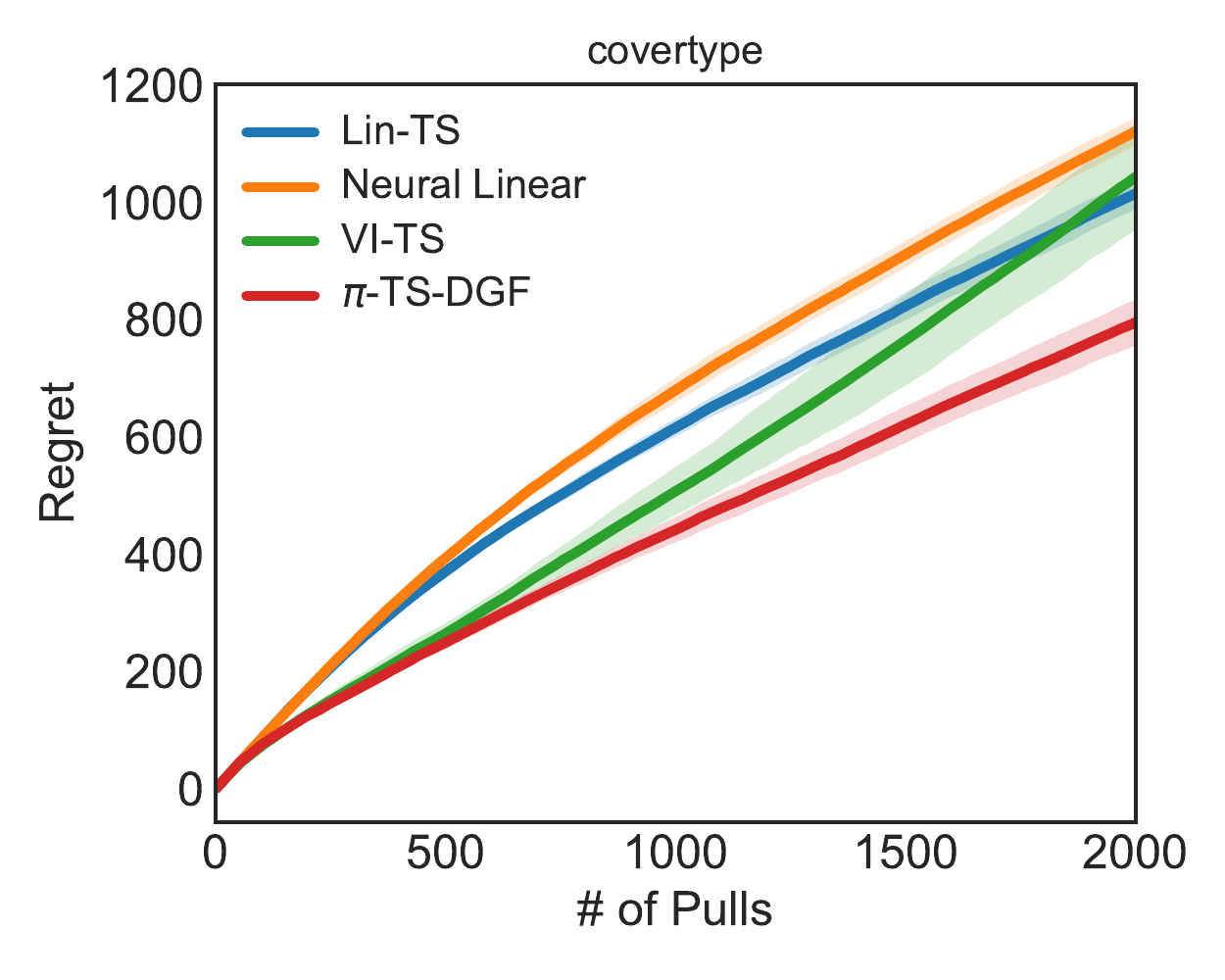}
	\end{tabular}
	\vspace{-5mm}
	\caption{Normalized Regret comparison on real-world datasets.}
	\label{fig:deep}
	\vspace{-4mm}
\end{figure}

\addtolength{\tabcolsep}{-1pt}
\begin{table*}[h!]
	\centering
	\caption{\label{tab:quantativeall} Normalized accumulative regret of all methods.}
	\vspace{5pt}
	\begin{tabular}{lcccccc}
		\toprule[1.2pt]
		& Statlog & Adult & Covertype & Census & Financial & Mushroom\\
		\midrule
		Lin-TS &18.62 $\pm$ 0.02
		&96.71 $\pm$ 0.09
		&59.32 $\pm$ 0.08
		&94.54 $\pm$ 0.28
		&10.48 $\pm$ 0.04
		&\textbf{13.22 $\pm$ 0.13} \\
		Neural Linear &13.96 $\pm$ 0.03
		&97.55 $\pm$ 0.06
		&65.51 $\pm$ 0.07
		&79.15 $\pm$ 0.32
		&17.76 $\pm$ 0.06
		&14.25 $\pm$ 0.12  \\
		VI-TS & 23.32 $\pm$ 0.09& 95.66 $\pm$ 0.08& 60.99 $\pm$ 0.25& 64.85 $\pm$ 0.91& 35.2 $\pm$ 0.31& 23.32 $\pm$ 0.16 \\
		$\pi$-TS-DGF &\textbf{5.37 $\pm$ 0.03}
		&\textbf{90.76 $\pm$ 0.11}
		&\textbf{46.48 $\pm$ 0.10}
		&\textbf{49.85 $\pm$ 0.52}
		&\textbf{8.23 $\pm$ 0.10}
		&14.29 $\pm$ 0.07  \\
		\bottomrule[1.2pt]
	\end{tabular}
	\vspace{-5pt}
\end{table*}
\addtolength{\tabcolsep}{1pt}

\subsection{Effects of Numbers of Particles}
\begin{wrapfigure}{R}{5.2cm}
	\vspace{-6mm}
	\centering
	\hspace{-2mm}\includegraphics[width=1\linewidth]{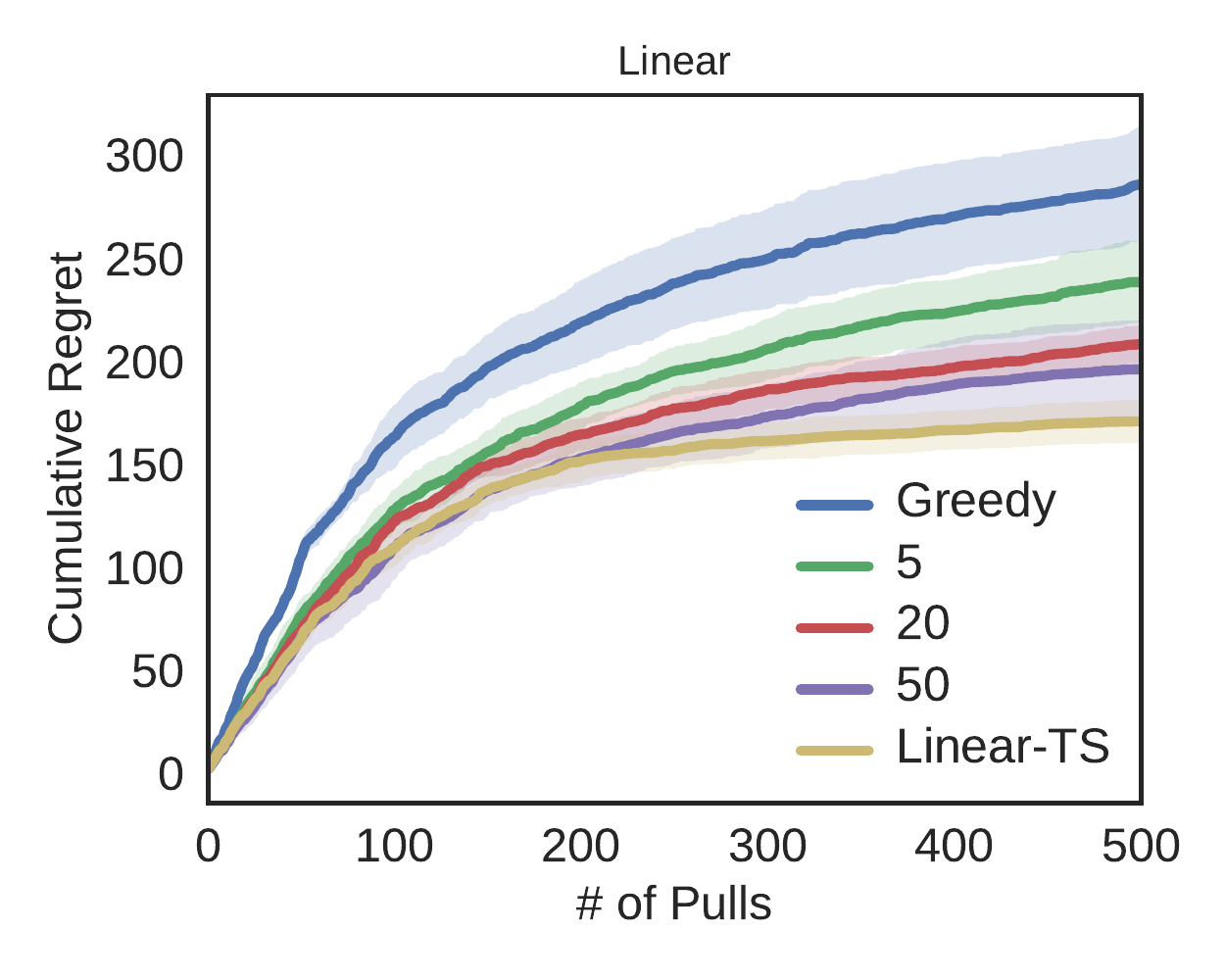}
	\vspace{-4mm}
	\caption{\small Impact of particle number.}
	\vspace{-7mm}
	\label{fig:parablation}
\end{wrapfigure}

We investigate
the influence of different number of particles used on
the performance. We choose different number of particles as $M=1,5,20, 50$, where the case of 1 particle corresponds to the greedy setting. We use the same model as the above experiments. In this part, we use larger noise $\sigma_i^2=0.1$, and pull each arm 2 times at the initial stage. Figure \ref{fig:parablation} shows accumulated regrets along with number of particles. As expected, the best performance is achieved with the largest number of particles. The performance keep improving with increasing particles, but the gain becomes insignificant considering the increased computational costs.

\begin{figure}[h]
	\centering
	\begin{tabular}{cc}
		\hspace{-4mm}
		\includegraphics[width=0.45\linewidth]{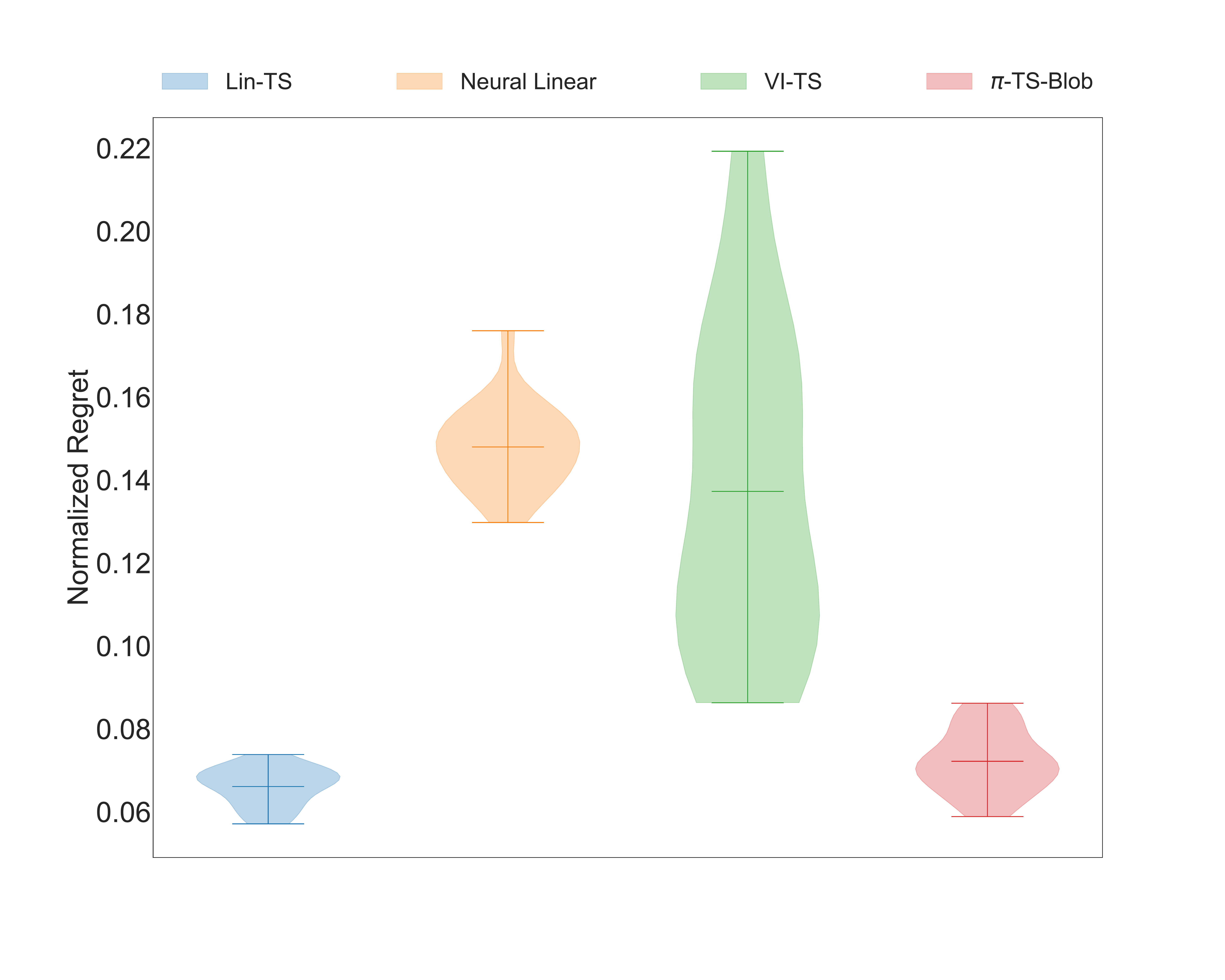}
		&   \hspace{-7mm}
		\includegraphics[width=0.45\linewidth]{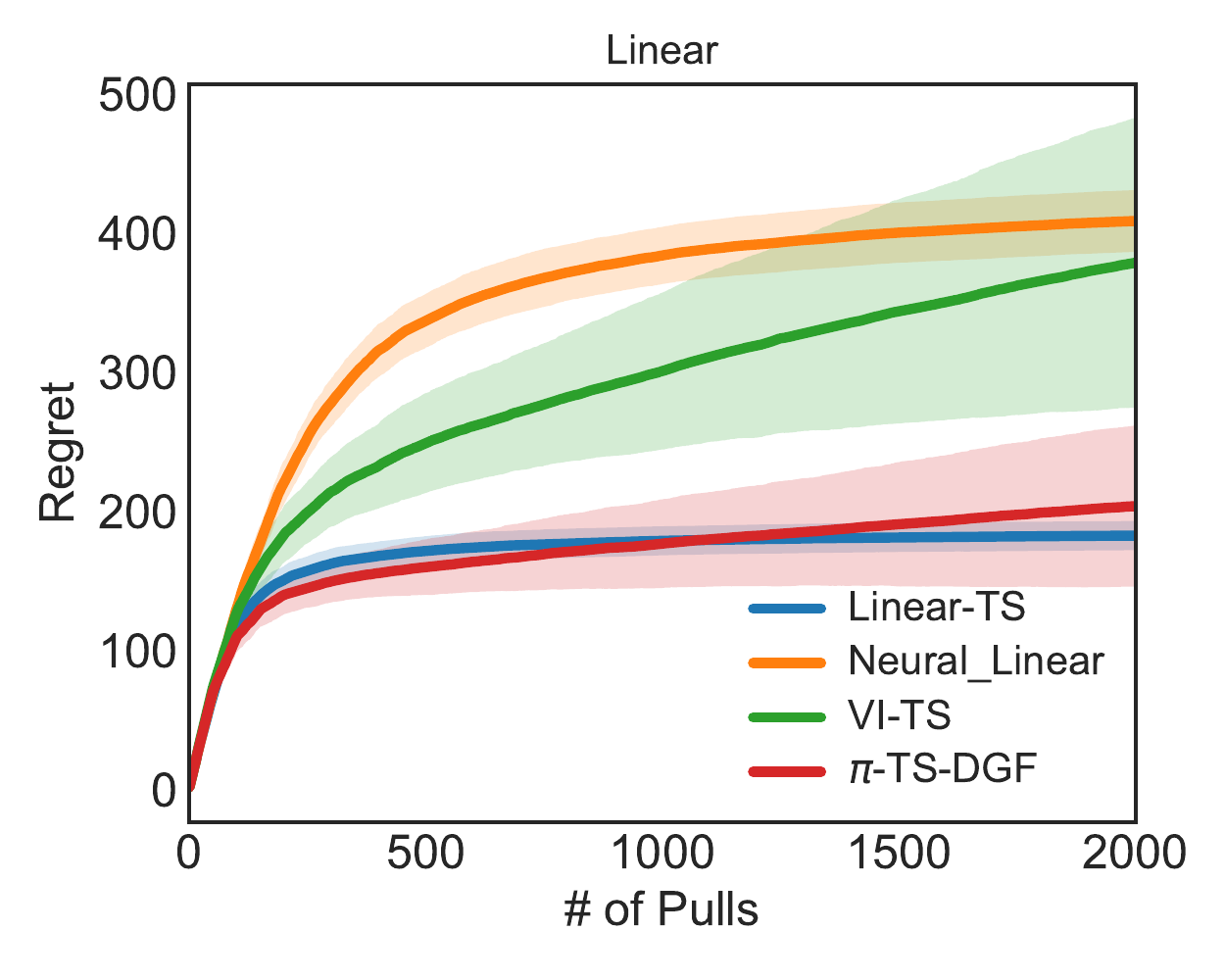}\\
	\end{tabular}
	\vspace{-3mm}
	\vspace{-2mm}
	\caption{ Normalized Regret comparison among four methods on linear cases.}
	\label{fig:linear}
	\vspace{-4mm}
\end{figure}

\subsection{Linear Rewards}
We consider a contextual bandit with $k=8$ arms and $d=10$ dimensional contexts. For a given context $X \sim \mathcal{N}(\mu, \Sigma)$, the reward obtained by pulling arm $i$ follows a linear model $r_{i, X} = X^T \beta_i + \epsilon$ with $\epsilon \sim \mathcal{N}(0, \sigma_i^2)$, where $\sigma_i^2=0.01i$
The posterior distribution over $\beta_i \in \mathrm{R}^d$ can be exactly computed using the standard Bayesian linear regression formula, denoted as Lin-TS. We set the contextual dimension $d = 20$, and the prior to be $\beta \sim \mathcal{N}(0, \lambda \ \text{I}_d)$, for  $\lambda > 0$. The results in terms of both regret and normalized regret are plotted in Figure \ref{fig:linear}. We can see from the figure that the proposed methods, $\pi$-TS-Blob and $\pi$-TS-DGF, perform almost as well as Lin-TS, the exact model; whereas other methods such as Neural-Linear and VI-TS receive much larger regrets. The gap is mostly caused by the approximation error between the exact posterior and approximate posterior. Especially, VI-TS shows a higher regret variance. Furthermore, both $\pi$-TS-Blob and $\pi$-TS-DGF are found performed similarly.

\subsection{Sparse Linear Rewards}
\begin{figure}[h!] \centering
	\vspace{-3mm}
	\begin{tabular}{cc}
		\hspace{-4mm}
		\includegraphics[width=0.45\linewidth]{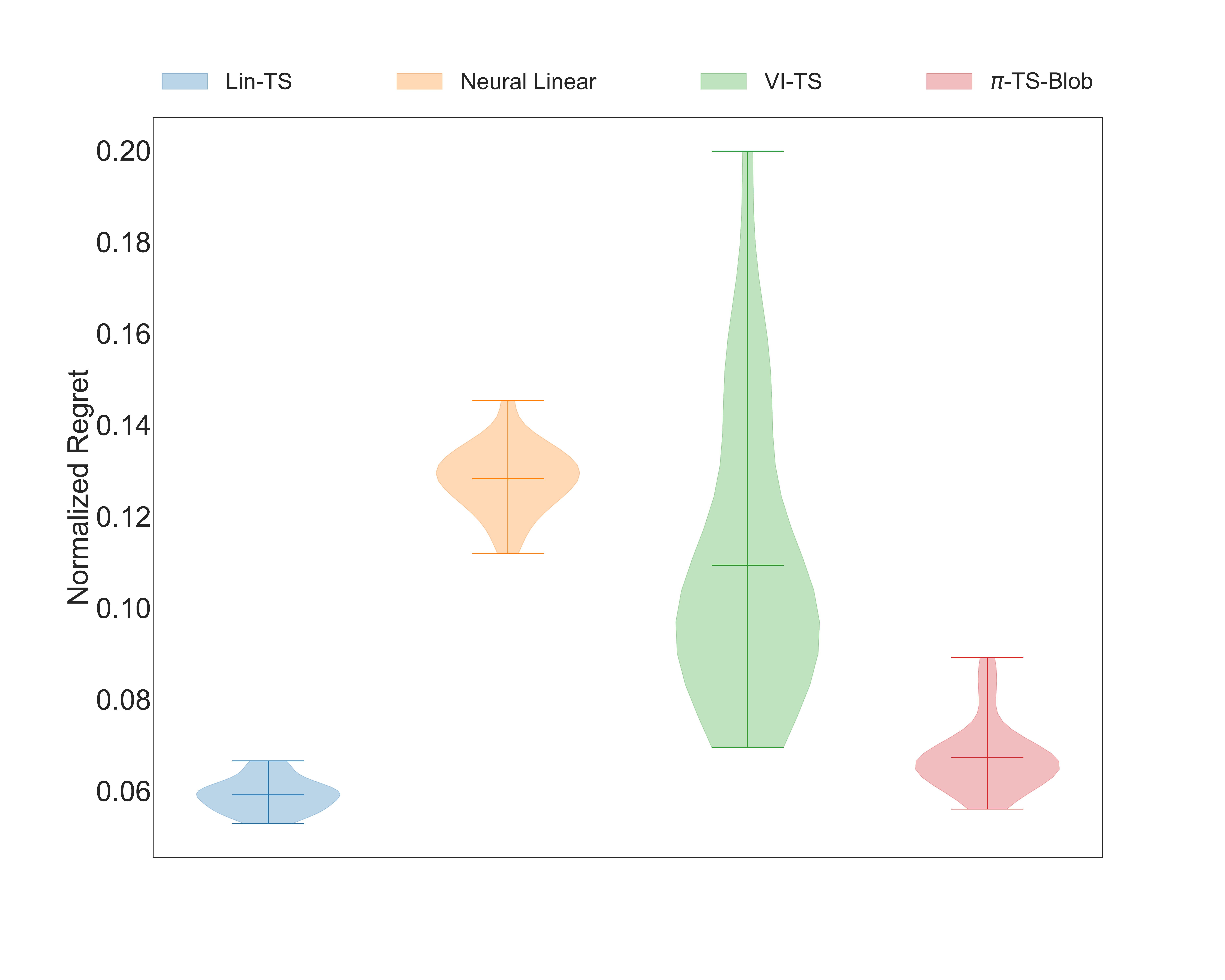}
		&   \hspace{-7mm}
		\includegraphics[width=0.45\linewidth]{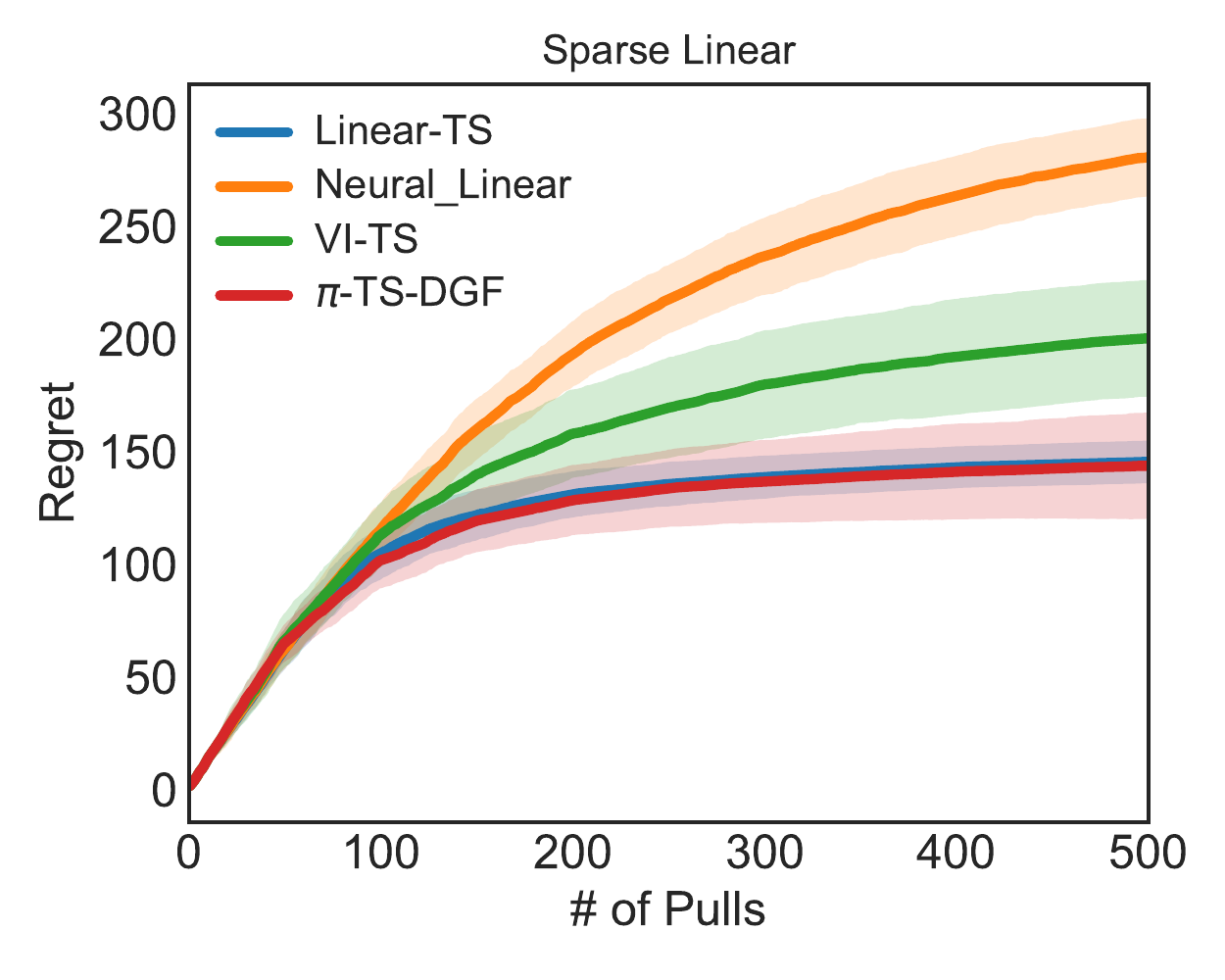}\\
	\end{tabular}
	\vspace{-3mm}
	\vspace{-2mm}
	\caption{ Normalized Regret comparison among five methods on sparse linear cases.}
	\label{fig:sparselinear}
\end{figure}

In this case, the weight vector $\beta^s_i \in \mathrm{R}^d$ is sparse. Specifically, $\beta^s_i$ is more sparse than the standard $\beta$ used above.
The reward obtained by pulling arm $i$ follows a sparse linear model is $r_{i, X} = X^T \beta^s_i + \epsilon$, where $\sigma_i^2=0.01i$. The results are plotted in Figure \ref{fig:sparselinear}. Similarly, much less regrets are achieved by $\pi$-TS-DGF, which is comparable to Lin-TS.

\section{More Details of Methods}
\subsection{Vanilla Thompson Sampling}
\begin{algorithm}
	\caption{Thompson Sampling (Vanilla Version)}
	\label{algo:vanilla}
	\begin{algorithmic}[1]
		\REQUIRE {prior distribution $p_0$}
		\FOR{$t = 1,2,\ldots,T$}
		\STATE Observe context $\xb_t$
		\STATE Draw $\hat{\thetab}_{t}$ from the posterior $p_{t-1}$ 
		\STATE Select $\ab_t \in \arg\max_{\ab} m(\xb_t, \ab; \hat{\thetab}_{t})$
		\STATE Observe and receive reward $r_t$
		\STATE Update posterior $p_t (\thetab) \leftarrow p_{t-1}(\thetab | (\xb_t, \ab_t, r_t))$.
		%
		\ENDFOR
	\end{algorithmic}
\end{algorithm}
\subsection{Gradient Flows in Euclidean space}
 For ease of understanding, we first motivate from gradient flows on the Euclidean space in the following. \label{sec:gf_euc}
For a smooth function\footnote{We will focus on the convex case, since this is the case for many gradient flows on the space of probability measures, as detailed subsequently.} $F: \mathbb{R}^d \rightarrow \mathbb{R}$, and a starting point $\thetab_0 \in \mathbb{R}^d$, the gradient flow of $F(\thetab)$ is defined as the solution of the differential equation: $\frac{\mathrm{d}\thetab}{\mathrm{d}\tau} = -\nabla F(\thetab(\tau))$, for time $\tau > 0$ and initial condition $\thetab(0) = \thetab_0$. This is a standard Cauchy problem \citep{rulla1996error}, endowed with a unique solution if $\nabla F$ is Lipschitz continuous. When $F$ is non-differentiable, the gradient is replaced with its subgradient, which gives a similar definition, omitted here for simplicity.

We consider a contextual bandit with $k=8$ arms and $d=10$ dimensional contexts.
For a given context $X \sim \mathcal{N}(\mu, \Sigma)$, the reward obtained by pulling arm $i$ follows a linear model $r_{i, X} = X^T \beta_i + \epsilon$ with $\epsilon \sim \mathcal{N}(0, \sigma_i^2)$, where $\sigma_i^2=0.01i$
The posterior distribution over $\beta_i \in \mathrm{R}^d$ can be exactly computed using the standard Bayesian linear regression formula, denoted as Lin-TS. We set the contextual dimension $d = 20$, and the prior to be $\beta \sim \mathcal{N}(0, \lambda \ \text{I}_d)$, for  $\lambda > 0$.
\vspace{-2mm}

\section{Related Work}\vspace{-0.1cm}
It is difficult in general to calculate exact posteriors in Thompson sampling. Thus it is necessary to efficiently approximate a posterior distribution to make TS scalable for complex models. \citep{blundell2015weight} first used standard variational inference to approximate the posterior of neural networks, {\it i.e.}, Bayesian neural networks, which were then incorporated into Thompson sampling. Further, \citep{osband2016deep} proposed to use different heads for a deep Q-network to approximate posterior with bootstrap. Inspired by \citep{osband2016deep}, \cite{lu2017ensemble} proposed ensemble sampling, which uses a set of particles to approximate a posterior distribution. These particles are updated independently with stochastic gradient descent, without a convergence guarantee, in terms of posterior-approximation convergence. Similarly, weighted bootstrap~\citep{vaswani2018new} uses random weights performed on the likelihood to mimic the bootstrap, which is connected to TS. \citep{riquelme2018deep} built a benchmark to evaluate deep Bayesian bandits, and especially recommended the neural linear method, which uses a deep neural network to extract features and perform linear Thompson sampling based on these features. Similar to neural linear, \citep{azizzadenesheli2018efficient} replaced the final layer of a deep neural network with Bayesian logistic regression for deep Q-networks, which greatly boosted the performance on Atari benchmarks. \citep{zhang2017learning} firstly investigate particle-based Thompson sampling in contextual bandits settings. \citep{zhang2018policy} places policy optimization into the space of probability measures, and interpret it as Wasserstein gradient flows.  In this work, we provide a distribution optimization perspective to understand the posterior approximation, and propose efficient algorithms to approximate posterior distributions in Thompson sampling. This work can be regarded as the counterpart of \citep{zhang2017learning} for value-based methods.
\end{document}